\documentclass[runningheads]{llncs}
\usepackage[T1]{fontenc}
\usepackage{graphicx}
\usepackage{graphicx}
\usepackage{amsmath}
\usepackage{booktabs} 
\usepackage{bbding} 
\usepackage{makecell} 
\usepackage{mathrsfs}
\usepackage[normalem]{ulem}
\useunder{\uline}{\ul}{}
\usepackage[inline]{enumitem}
\usepackage{enumitem}
\usepackage{xcolor}
\definecolor{darkgreen}{rgb}{0.0, 0.5, 0.0}
\usepackage{makecell}
\usepackage{CJK}
\usepackage{capt-of}
\usepackage{hyperref}

\begin{document}
\title{Modeling Comparative Logical Relation with Contrastive Learning for Text Generation}
\titlerunning{Modeling Comparative Logical Relation with Contrastive Learning}

\author{Yuhao Dan\inst{1,2,3} \and
Junfeng Tian\inst{4} \and
Jie Zhou\inst{1,2,3} \and
Ming Yan\inst{5} \and
Ji Zhang\inst{5} \and
\mbox{Qin Chen}\inst{1,2,3} \thanks{Corresponding author: qchen@cs.ecnu.edu.cn} \and
Liang He\inst{1,2,3}}
\authorrunning{Y. Dan et al.}

\institute{Lab of Artificial Intelligence for Education, East China Normal University \and
Shanghai Institute of Artificial Intelligence for Education, ECNU \and
School of Computer Science and Technology, ECNU \and
Xiaohongshu Inc \and
Alibaba Group 
}

\maketitle              
\vspace{-5mm}
\begin{abstract}
Data-to-Text Generation (D2T), a classic natural language generation problem, aims at producing fluent descriptions for structured input data, such as a table. Existing D2T works mainly focus on describing the superficial \textit{associative relations} among entities, while ignoring the deep \textit{comparative logical relations}, such as A is better than B in a certain aspect with a corresponding opinion, which is quite common in our daily life. In this paper, we introduce a new D2T task named comparative logical relation generation (CLRG). Additionally, we propose a \textbf{Co}mparative \textbf{Lo}gic (\textsc{CoLo}) based text generation method, which generates texts following specific \textit{comparative logical relations} with contrastive learning. Specifically, we first construct various positive and negative samples by fine-grained perturbations in entities, aspects and opinions. Then, we perform contrastive learning in the encoder layer to have a better understanding of the \textit{comparative logical relations}, and integrate it in the decoder layer to guide the model to correctly generate the relations. Noting the data scarcity problem, we construct a Chinese Comparative Logical Relation Dataset (CLRD), which is a high-quality human-annotated dataset and challenging for text generation with descriptions of multiple entities and annotations on their \textit{comparative logical relations}. Extensive experiments show that our method achieves impressive performance in both automatic and human evaluations. 

\vspace{-1mm}
\keywords{Natural language processing \and Data-to-text generation \and Contrastive learning \and Dataset construction.}
\end{abstract}
\vspace{-7mm}
\section{Introduction}
\begin{figure}[t]
\vspace{-2mm}
\begin{center}
\includegraphics[width=1\textwidth]{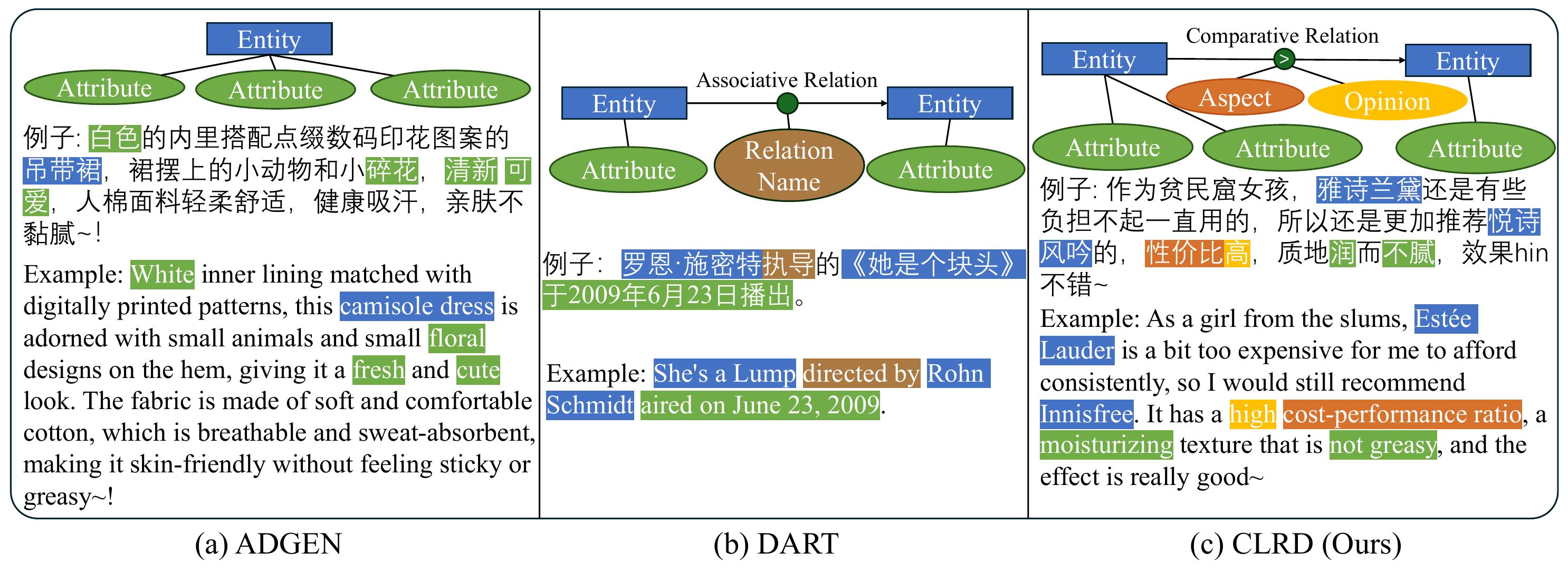}
\end{center}
\vspace{-6mm}
\caption{Examples from two popular Data-to-Text datasets and the proposed Chinese Comparative Logical Relation Dataset (CLRD): (a) Descriptions based on a single entity with attributes; (b) Descriptions with an \textit{associative relation} (e.g., "directed by"); (c) Descriptions with a \textit{comparative logical relation}, including aspects, opinions, and order. Entities and relations are color-coded.} 
\vspace{-4mm}
\label{intro_figure}
\end{figure}

\vspace{-1mm}
Data-to-Text Generation (D2T) is a long-established task in natural language processing (NLP) that aims to convert structured data, like tables and keywords, into natural language \cite{DBLP:journals/csl/DusekNR20,DBLP:conf/aaai/ZhanZCSDBYL21}. Existing works primarily focus on verbalizing single entities with multiple attributes, as demonstrated in Figure \ref{intro_figure}(a). \cite{shao-etal-2019-long,DBLP:conf/aaai/ZhangZZZDCDHHXL22,DBLP:conf/kdd/0002Z0XLW22,yuan2024screening}.
Recently, research interest in D2T has shifted towards modeling \textit{associative relations} (e.g., is part of, directed by) between entities, as shown in Figure \ref{intro_figure}(b) \cite{gardent-etal-2017-creating,nan-etal-2021-dart}. 
These relations can be derived from knowledge graphs or syntactic structures, reflecting the associations between two entities. 

In addition to associative relations, \textit{comparative logical relations} (CLRs) are also crucial for humans. Current studies have found that making comparisons enhances decision making \cite{hsee1998will}, improves learning outcomes \cite{gentner1997structure} and boosts social understanding \cite{silver2010compare}. As illustrated in \ref{intro_figure}(c), a CLR between two entities can be formalized as: entity A is better than entity B in a certain aspect with an opinion.
Specifically, in the example, the entity A is ``Innisfree", which is considered ``higher" than ``Estée Lauder" in terms of the ``cost-performance ratio".

Despite the importance of CLRs for humans, few studies have explored how well machines can emulate this capability. Generating text with CLRs faces three main challenges. First, maintaining fluency and coherence when describing relations with multiple comparative elements (e.g., entities, attributes, aspects, and opinions) is a general challenge in text generation. Second, the inclusion of comparative aspects and opinions makes it more difficult to cover all necessary comparative elements. Lastly, verbalizing CLRs requires more than just surface-level articulation of comparative elements; it demands a profound and authentic understanding of comparative logic. For example, when describing the relation in \ref{intro_figure}(c), the model must maintain the correct comparative order of ``Innisfree" and ``Estée Lauder".

To tackle the aforementioned challenges, we propose the CoLo method to model comparative logical relations (CLRs) using contrastive learning for text generation. First, we create positive and negative samples through fine-grained perturbations in comparative elements. Specifically, synonym replacement generates positive samples to facilitate learning of alias variants, while entity swapping, aspect substitution, and opinion substitution produce negative samples, enhancing the model’s ability to handle various comparative elements accurately. We then implement a two-stage contrastive learning strategy to enhance text generation with CLRs. This approach improves the model's understanding of CLRs through contrastive encoding and ensures the generated output adheres to the input CLRs via contrastive decoding.
We compare our method with the advanced D2T models and GPT-3.5, and conduct automatic and human evaluations to verify the effectiveness. 

The main contributions of our work are: (1) We introduce a Comparative Logical Relation Generation task with a new dataset, which advances research in text generation involving intricate logical relations; (2) We propose a novel method to model the \textit{comparative logical relations} with two-staged contrastive learning for text generation, where the contrastive encoding facilitates the understanding of the relations and the contrastive decoding encourages text generation with correct comparative logic; (3) We conduct extensive experiments with in-depth analyses, and the results have verified the superiority of our proposed method in verbalizing \textit{comparative logical relations} using only 0.58B parameters.

\begin{table}[t!]
\vspace{-2mm}
\caption{Statistics of CLRD and other data-to-text datasets. \textbf{CLR} represents \textit{comparative logical relation}.}
\label{tab:related_datasets}
\vspace{-2mm}
\centering
\small
\setlength{\tabcolsep}{1mm}{\begin{tabular}{lllccll}
\toprule
\textbf{Name} & \textbf{Size} & \textbf{Length} & \makecell[l]{\textbf{MultiEntity}}& \makecell[c]{\textbf{CLR}} & \makecell[c]{\textbf{Source}} & \textbf{Language} 
\\ \midrule  WebNLG\cite{gardent-etal-2017-creating} & 39K/17K & 20/17   & \XSolidBrush &  \XSolidBrush  & DBpedia  & en/ru   \\ 
                E2E\cite{DBLP:journals/csl/DusekNR20} & 13K  &  21  & \XSolidBrush &  \XSolidBrush  & restaurant    &  en            \\ 
              ToTTo\cite{parikh-etal-2020-totto} & 137K & 14 & \XSolidBrush & \XSolidBrush & Wikipedia & en \\ 
              ADGEN\cite{shao-etal-2019-long} & 119K & 110 & \XSolidBrush & \XSolidBrush & e-commerce & cn \\ 
              JDPDG\cite{DBLP:conf/aaai/ZhanZCSDBYL21} & 346K & 40 & \XSolidBrush & \XSolidBrush & e-commerce & cn \\ 
               FPDG\cite{chan-etal-2019-stick} & 414K & 63 & \XSolidBrush & \XSolidBrush & e-commerce & cn \\ 
              CLRD & 15K & 120  &\Checkmark & \Checkmark & e-commerce& cn \\ \bottomrule    
\end{tabular}}
\vspace{-4mm}
\end{table}

\vspace{-2mm}
\section{Dataset Construction}
\vspace{-2mm}
\subsection{Data Collection and Annotation}
\label{sec:dcaa}
\vspace{-1mm}

Due to the lack of specific datasets, we utilize the comprehensive CommonCrawl (CC) dataset to extract e-commerce beauty product reviews. Our preliminary analysis revealed that these texts contain rich comparative logical relations (CLRs). To annotate the CLRs accurately and efficiently, we recruited four annotation experts from Alibaba iTag platform for each data point. The annotators were tasked with identifying and extracting consecutive sentences describing a CLR and highlighting the comparative elements within that span.

Since the entities in the comparative elements are all beauty products, we also annotated the following six categories of attributes for each entity: brand name, ingredient, efficacy, texture, appearance, and fragrance. On average, each category contains 165 distinct attributes after this process. To ensure the quality of the annotations, we evaluated inter-annotator agreement (IAA) using Krippendorff's alpha coefficient and achieved an average score of 0.83, indicating a strong consensus among the annotators.

The Comparative Logical Relation Dataset (CLRD) contains 15,104 data points, with CLR narrations averaging around 120 words. We split the dataset into training, validation, and test sets in an 8:1:1 ratio. A sample data demonstration is provided in Table \ref{tab:case_study}.

\vspace{-2mm}
\subsection{Comparison with other Datasets}
\label{subsec: comparative logical relation dataset}
\vspace{-1mm}



We compare our CLRD with existing D2T datasets (See Table \ref{tab:related_datasets}). 
We observe that all other datasets only contain a single entity and lack annotations of comparative logical relations (CLRs).
In contrast, our dataset not only includes comparisons of two entities but also provides explicit CLR annotations, which can significantly enhance research in logically coherent text generation. Additionally, our dataset has the longest average length, making the CLRG task more challenging.
While some datasets are larger in size, they are often created automatically and may introduce noise. In comparison, our dataset is meticulously curated and thoroughly reviewed by humans, ensuring high quality.

\begin{figure*}[t]
\vspace{-3mm}
\begin{center}
\includegraphics[width=1\textwidth]{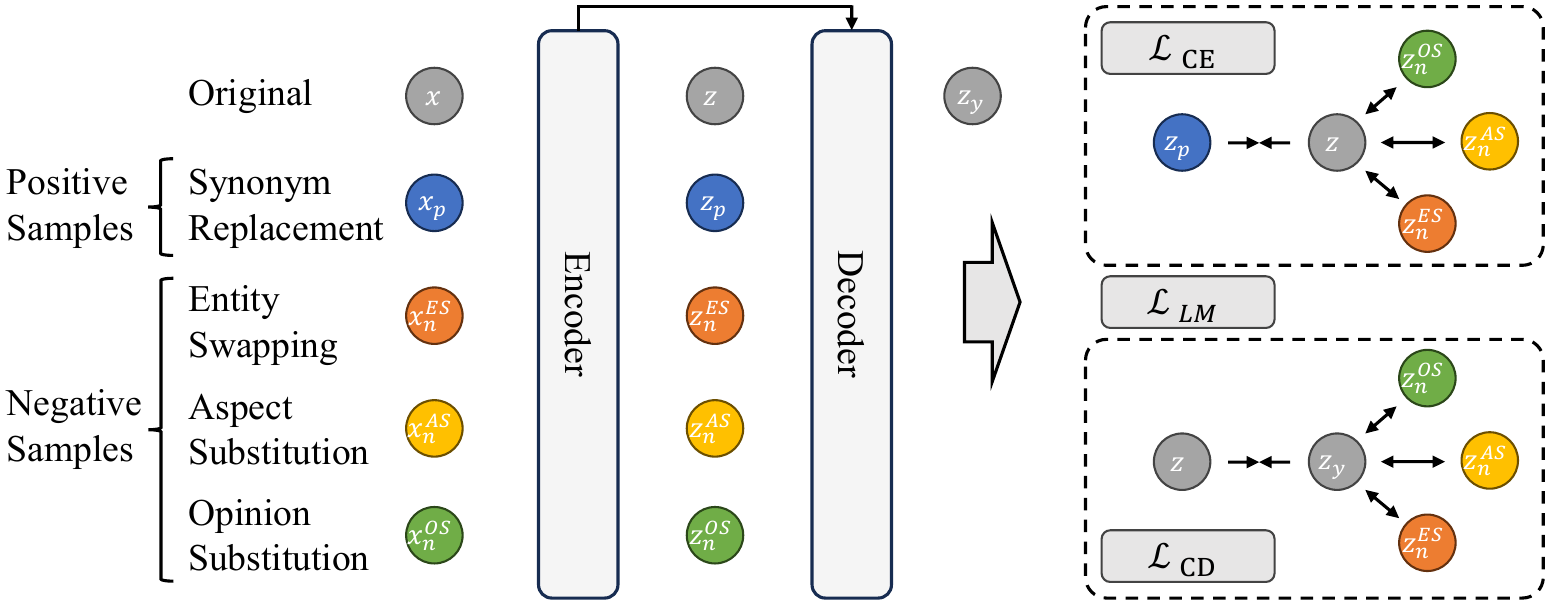}
\end{center}
\vspace{-6mm}
\caption{
Overview of our proposed method. On the left, our strategy for constructing contrastive samples is shown. On the right, we introduce two contrastive losses beyond the standard language modeling loss ($\mathcal{L}_{\mathrm{LM}}$): the contrastive encoding loss ($\mathcal{L}_{\mathrm{CE}}$) and the contrastive decoding loss ($\mathcal{L}_{\mathrm{CD}}$).}
\label{method_figure}
\vspace{-4mm}
\end{figure*}

\vspace{-2mm}
\section{Method}
\vspace{-1mm}
The overview of our proposed method is shown in Figure \ref{method_figure}. 
First, we construct positive and negative samples with multiple strategies for further contrastive learning. Then, we present two modules within the encoder-decoder framework, namely Contrastive Encoding and Contrastive Decoding, which incorporate the contrastive encoding loss ($\mathcal{L}_{\mathrm{CE}}$) and the contrastive decoding loss ($\mathcal{L}_{\mathrm{CD}}$) besides the general language modeling loss ($\mathcal{L}_{\mathrm{LM}}$) for text generation. The $\mathcal{L}_{\mathrm{CE}}$ focuses on having a better understanding of the \textit{comparative logical relations} 
 (CLRs) in the encoding stage, while the $\mathcal{L}_{\mathrm{CD}}$ ensures that the generated texts adhere to the input relations during the decoding stage.

\vspace{-2mm}
\subsection{Problem Formulation}
\label{sec:pf}
\vspace{-1mm}
In this paper, we propose the Comparative Logical Relation Generation (CLRG) Task, which aims to verbalize comparative logical relations (CLRs). 
For clarity, the input CLR can be represented as a tuple $x=\left (e_a, e_b, a, o \right )$, where the comparative elements inside denote Entity A, Entity B, Aspect, and Opinion, respectively.
The model output $y$ should clearly express that $e_a$ is better than $e_b$ in aspect $a$ based on opinion $o$.
For example, in Table \ref{tab:case_study}, $x$ can be the CLR between two beauty products, and $y$ can be a customer review describing $x$.



\vspace{-2mm}
\subsection{Contrastive Sample Construction}
\label{sec:cec}
\vspace{-1mm}
It is essential for the model to mention both compared entities $e_a$ and $e_b$ in the output description while maintaining the correct comparative order. Besides, the model should correctly identify the aspect $a$ being compared and the corresponding opinion $o$.

\begin{figure}[t!]
\vspace{-2mm}
\begin{center}
\includegraphics[width=1.0\textwidth]{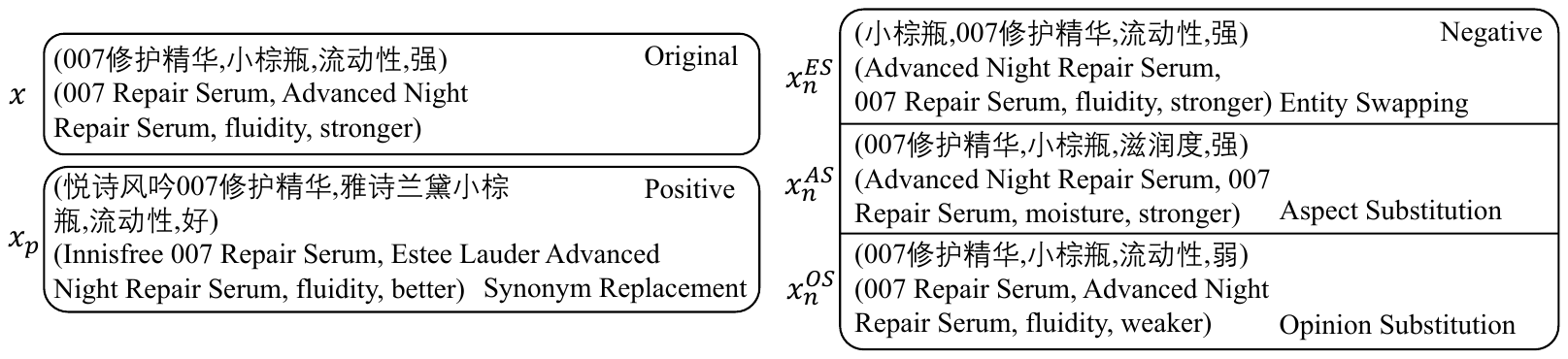}
\end{center}
\vspace{-6mm}
\caption{Examples of an original tuple and its contrastive tuples. 
The positive tuple is derived by replacing words in the original tuple with their synonyms. The negative tuples are constructed by entity swapping, aspect substitution and opinion substitution. Details can be found in Section \ref{sec:cec}.
}
\vspace{-4mm}
\label{contrastive_samples}
\end{figure}

To meet these challenges, we create contrastive samples for the original tuple $x$ (see Figure \ref{contrastive_samples}) to expose the model to various aliases of the same entity, aspect, and opinion. We generate a positive example $x_p$ by replacing words in $x$ with their synonyms. Furthermore, to address the insensitivity of existing models to the order of entities, the comparative aspects and the opinions, we construct negative examples using three different approaches, thereby capturing the nuances of \textit{comparative logical relations} (CLRs). (1) Entity Swapping (ES): we swap the order of two entities in $x$ and construct $x_n^\mathrm{ES}=\left(e_b,e_a,a,o\right)$; (2) Aspect Substitution (AS): we substitute the aspect with a random aspect $a^{\prime}$ and obtain $x_n^\mathrm{AS}=\left(e_a,e_b,a^{\prime},o\right)$; (3) Opinion Substitution (OS): we replace the opinion with one of its antonyms (if it exists) or a random opinion $o^{\prime}$ and obtain $x_n^\mathrm{OS}=\left(e_a,e_b,a,o^{\prime}\right)$.

\vspace{-2mm}
\subsection{Contrastive Encoding}
\vspace{-1mm}

\label{sec:CRML}
One limitation of most pre-trained generation models is their lack of sensitivity to CLRs, which plays a significant role in Comparative Logical Relation Generation. For instance, when the order of entities in a tuple is switched, it completely changes the meaning of the CLR. However, existing encoder models may generate similar embeddings for these tuples, failing to capture this crucial distinction.

To solve the problem, we propose a Contrastive Encoding (${\mathrm{CE}}$) strategy to model the relations among entities, aspects and opinions by maximizing the similarity between positive pairs of samples and minimizing the negatively associated pairs.
Given the original tuple or its contrastive tuples (e.g., $x$, $x_p$, $x_{n}^\mathrm{ES}$, $x_{n}^\mathrm{AS}$, $x_{n}^\mathrm{OS}$), we take the mean pooling of encoder outputs as representations of corresponding CLRs (e.g., $z$, $z_p$, $z_{n}^\mathrm{ES}$, $z_{n}^\mathrm{AS}$, $z_{n}^\mathrm{OS}$).
We calculate the distance between two representations with score function $s(\cdot, \cdot)$, which measures the cosine similarity. To measure the distance between the original tuple with its positive and negative tuples, we create two sets of similarity scores, $\mathcal{P}^{+}$ for positive pairs and $\mathcal{P}^{-}$ for negative pairs. 
We optimize the margin among derived pairs with the following loss:
\begin{equation}
\small
    \begin{split}
    \mathcal{L}_{\mathrm{CE}}
    = \sum_{p^{+}\in \mathcal{P}^{+}} 
    \sum_{p^{-}\in \mathcal{P}^{-}} \max \left\{0,  p^{-} - p^{+} + \xi \right\} 
    \label{equa:CRML}
    \end{split}
\end{equation}
In Equation \ref{equa:CRML}, $\mathcal{P}^{+}=\left \{s(z,z_{p}) \right \}$, and  $\mathcal{P}^{-} =\left \{s(z,{z_{n}^{\mathrm{ES}}}), s(z,{z_{n}^{\mathrm{AS}}}), s(z,{z_{n}^{\mathrm{OS}}}) \right \}$. 
Treating all negative tuples in $\mathcal{P}^{-}$ equally with a fixed value for $\xi$ in Equation \ref{equa:CRML} is not appropriate. The negative tuples that are easier to generate the target sentence should be punished more. This can be quantified using a sentence-level metric (e.g., generation loss of a negative tuple deriving the target sentence). Therefore, we set $\xi_{i}=\gamma \ast f_{r}(\mathcal{L}_{\mathrm{LM}}(z_{n}^{i}))$, where $i \in \left \{ \mathrm{ES}, \mathrm{AS}, \mathrm{OS} \right \}$. The function $f_{r}$ ranks the language modeling loss for each negative tuple deriving the target sentence, in descending order of their values. For example, $f_{r} \left (0.56, 0.87, 0.24 \right )= \left (2, 1, 3 \right )$. This reflects the difference in generating the target sentence's possibility, where $\gamma$ is an adjustable hyperparameter controlling the strength.
\vspace{-2mm}
\subsection{Contrastive Decoding}
\label{sec:RDCL}
To let the model learn the semantics distances between encoded \textit{comparative logical relations} and decoded descriptions, we propose Contrastive Decoding (${\mathrm{CD}}$) strategy. 
As we did in Section \ref{sec:CRML}, we calculate the mean pooling of decoder outputs as representations for the output descriptions for the original example (e.g., $z_y$).
Since the encoder outputs and the decoder outputs are not at the same semantic level, we use two Fully Connected Neural Networks to transform them prior to assessing their similarity. We still use $s(\cdot, \cdot)$ mentioned in Section \ref{sec:CRML} as the similarity metric. The definition of $\mathcal{L}_\mathrm{CD}$ follows the same form as Equation \ref{equa:CRML}, with the exception that $\mathcal{P}^{+} =\left \{s(z_y,z) \right \}$ and $\mathcal{P}^{-} =\left \{s(z_y,{z_{n}^{\mathrm{ES}}}), s(z_y,{z_{n}^{\mathrm{AS}}}), s(z_y,{z_{n}^{\mathrm{OS}}}) \right \}$.
During the training phase, we use the objective: $\mathcal{L}=\mathcal{L}_{\mathrm{LM}}+\mathcal{L}_{\mathrm{CE}}+\mathcal{L}_{\mathrm{CD}}$, where $\mathcal{L}_{\mathrm{LM}}$ is the next-token prediction loss.

\vspace{-2mm}
\section{Experimental Setup}
\subsection{Evaluation Metrics}
\label{sec:metrics}
\textbf{Automatic Evaluation Metrics.} We assess the fluency of generated texts using perplexity (\textbf{PPL}) based on a pre-trained Chinese language model, as per previous work. To measure the overall quality of the generated text, we employ several metrics: BLEU (\textbf{B-1}, \textbf{B-4}), ROUGE-L (\textbf{R-L}), \textbf{METEOR}, Distinct-4 (\textbf{Dist-4}), and \textbf{BERTScore}. Given the complexity of the input tuple components $(e_a, e_b, a, o)$, we use coverage (\textbf{Cover}) for evaluation to ensure the generated text includes all input components. However, high coverage alone does not guarantee the correct logical sequence of the \textit{comparative logical relation} (CLR). Therefore, we use the entailment score (\textbf{Entail}) \cite{bowman-etal-2015-large}, which is widely used to determine logical entailment between a premise and a hypothesis. For evaluation, the input tuple is verbalized into a sentence and treated as the hypothesis, while the generated text serves as the premise. If the input tuple can be inferred from the generated text, the logical relation is considered correct, indicating that the text follows the correct CLR.
We implement the entailment score using a multilingual BERT model\footnote{\href{https://huggingface.co/bert-base-multilingual-uncased}{https://huggingface.co/bert-base-multilingual-uncased}}, fine-tuned with a next sentence prediction objective on the XNLI dataset \cite{conneau2018xnli} to acquire basic NLI capabilities, and further trained on the CLRD training set to develop the ability to judge \textit{comparative logic}.

\noindent\textbf{Human Evaluation Metrics.}
In addition to the automatic evaluation metrics, we conduct human evaluation with three native Chinese-speaking annotators. The assessment is based on the following criteria:
\textbf{Fluency} reflects the clarity and comprehensibility of the generated description.
\textbf{Entity} assesses the incorporation of the input entities in the output.
\textbf{Aspect} evaluates the inclusion of the input comparative aspects in the output.
\textbf{Relation} measures the accuracy of describing the given input CLRs.
\textbf{Overall} is determined by the aforementioned criteria and the general quality of the output.
Each criteria is evaluated on a 0 to 3 scale, which is subsequently re-scaled to 0 to 100 for clarity. 

\vspace{-2mm}
\subsection{Baselines}
\vspace{-1mm}
We compare our method with the recent advanced baselines:
\textbf{1) BART}  \cite{lewis-etal-2020-bart} is a sequece-to-sequence language model pretrained on English data. We use a Chinese version BART\footnote{\href{https://huggingface.co/fnlp/bart-base-chinese}{https://huggingface.co/fnlp/bart-base-chinese\label{zh_bart}}} in experiments.
\textbf{2) mT5} \footnote{\href{https://huggingface.co/google/mt5-base}{https://huggingface.co/google/mt5-base}} \cite{xue-etal-2021-mt5} is a multilingual T5 \cite{DBLP:journals/corr/abs-1910-10683} model pre-trained on a large-scale dataset spanning 101 languages.
\textbf{3) Control Prefixes (ControlP)} \cite{Clive2021ControlPF} is the state-of-the-art model across several English D2T datasets such as WebNLG \cite{gardent-etal-2017-creating}, E2E \cite{DBLP:journals/csl/DusekNR20}, and DART \cite{nan-etal-2021-dart}.
\textbf{4) GPT-3.5}\footnote{The model we used was gpt-3.5-turbo-0613} is a large-scale language model developed by OpenAI, which is pre-trained on massive data and achieves good performance in various tasks.

\vspace{-2mm}
\subsection{Experimental Settings}
\vspace{-1mm}
We calculate perplexity based on a pre-trained Chinese language model \cite{zhang2021cpm} following previous work \cite{li-etal-2022-nominal}. 
We employ an identical mT5 model as the backbone for our method, along with ControlP, to ensure a fair comparison.
The value of $\gamma$ is set to 0.01 by searching from $\left [0.1, 0.01, 0.001 \right ]$. All trainable models are trained on the CLRD training set using the Adam optimizer with a learning rate searched from $\left [2e^{-3}, 2e^{-4}, 2e^{-5} \right]$. During decoding, the beam search size is set to 5 for all models. For GPT-3.5, we create a task-specific prompt and have the model generate descriptions given \textit{comparative logical relations}.

\vspace{-2mm}
\section{Results and Analyses}
\begin{table*}[t]
\vspace{-2mm}
\caption{Results of automatic evaluation. The best results are bolded, and the runners-up are underlined.} 
\label{tab:main_results}
\vspace{-2mm}
\centering
\small
\setlength{\tabcolsep}{1pt} 
\begin{tabular}{lcccccccccc}
\toprule
  & Param.& PPL$\downarrow$ & B-1 & B-4 & R-L & METEOR & Dist-4 & BERTScore & Entail & Cover \\
\midrule
BART &0.08B& 1.62 & 13.92 & 1.32 & 13.85 & 15.22 & \underline{93.89} & 62.72 & 37.06 & 36.18  \\
mT5 &0.58B& \underline{1.54} & \underline{20.91} & \underline{6.60} & \underline{21.85} & \underline{23.67} & 91.22 & 66.46 & 48.24 & 60.29  \\
ControlP &0.58B& 1.60 & 18.07 & 5.09 & 18.74 & 19.36 & 89.59 & 65.40 & 37.65 & 50.59 \\
GPT-3.5 &-& \textbf{1.43} & 11.72 & 1.61 & 10.79 & 14.51 & \textbf{94.00} & 61.17 & \textbf{63.53} & \textbf{69.71}\\
CoLo &0.58B& 1.55 & \textbf{22.41} & \textbf{7.51} & \textbf{22.00} & \textbf{25.18} & 92.43 & \textbf{67.76} & \underline{55.88} & \underline{63.97}  \\
\bottomrule
\end{tabular}
\vspace{-2mm}
\end{table*}

\begin{table}[ht!]
\caption{Results of human evaluation. The top results are bolded, and the runners-up are underlined.}
\label{tab:human_evaluation}
\vspace{-2mm}
\centering
\setlength{\tabcolsep}{1mm}{
\begin{tabular}{lcccccc}
\toprule
\textbf{}   &Param.& Fluency     & Entity     & Aspect     & Relation   & Overall  \\ \midrule
BART        &0.08B&87.84 (.42)&71.76 (.82)&61.96 (.81)&44.31 (.78)&40.20 (.70)\\
mT5         &0.58B&\underline{94.12} (.48)&85.29 (.87)&77.25 (.72)&60.78 (.77)&52.16 (.54)\\
ControlP    &0.58B&91.76 (.45)&76.47 (.84)&68.24 (.62)&52.55 (.69)&45.88 (.57)\\
GPT-3.5     &-&90.59 (.72)&\underline{88.24} (.86)&\textbf{92.94} (.77)& \textbf{76.86} (.65)& \textbf{68.24} (.56)\\ \hline
CoLo      &0.58B&\textbf{95.49} (.52)&\textbf{89.12} (.81)&\underline{80.00} (.77)&\underline{65.49} (.72)&\underline{56.86} (.62)\\ \bottomrule
\end{tabular}}
\vspace{-5mm}
\end{table}
\vspace{-2mm}
\subsection{Main Results}
To evaluate the effectiveness of our method, we conduct both automatic and human evaluations on the test set of CLRD, and the results are shown in Table \ref{tab:main_results} and Table \ref{tab:human_evaluation}. 

\noindent\textbf{Automatic Evaluation.}
We observe that our method outperforms all the baselines except GPT-3.5 in most cases. In particular, the improvement is more significant regarding to the Entail and Cover, indicating the superiority of our method in generating correct \textit{comparative logical relations}. We also notice that GPT-3.5 does not perform well on B-1, B-4, R-L, METEOR, and BERTScore, possibly due to its inability to be fine-tuned on downstream data, leading to low similarity with target sentences. In contrast, it excels in Entail and Cover, demonstrating its strong ability to follow instructions and generate texts with accurate logical relations. Notably, Colo achieves more than 85\% of the performance of GPT-3.5 regarding to the effective metrics as Entail and Cover with only 0.58B parameters. Moreover, our model is trainable, allowing easy adaptation to downstream data.

\noindent\textbf{Human Evaluation.}
 Each sample is judged by three annotators as described in Section \ref{sec:metrics}. We utilize the Krippendorff's alpha coefficient as a metric for assessing inter-annotator agreement. The resulting average score is 0.68, indicating a substantial level of consensus among the annotators. As shown in Table \ref{tab:main_results}, we achieve great improvements over the recent advanced baselines, such as BART, mT5 and ControlP. In addition, we obtain about 83.32\% of the overall performance of GPT-3.5, indicating the potential of our method in generating high-quality descriptions with \textit{comparative logical relations} using less parameters.

\vspace{-2mm}
\subsection{Ablation Studies}

\begin{figure*}[t]
\vspace{-4mm}
\begin{center}
\includegraphics[width=1.0\textwidth]{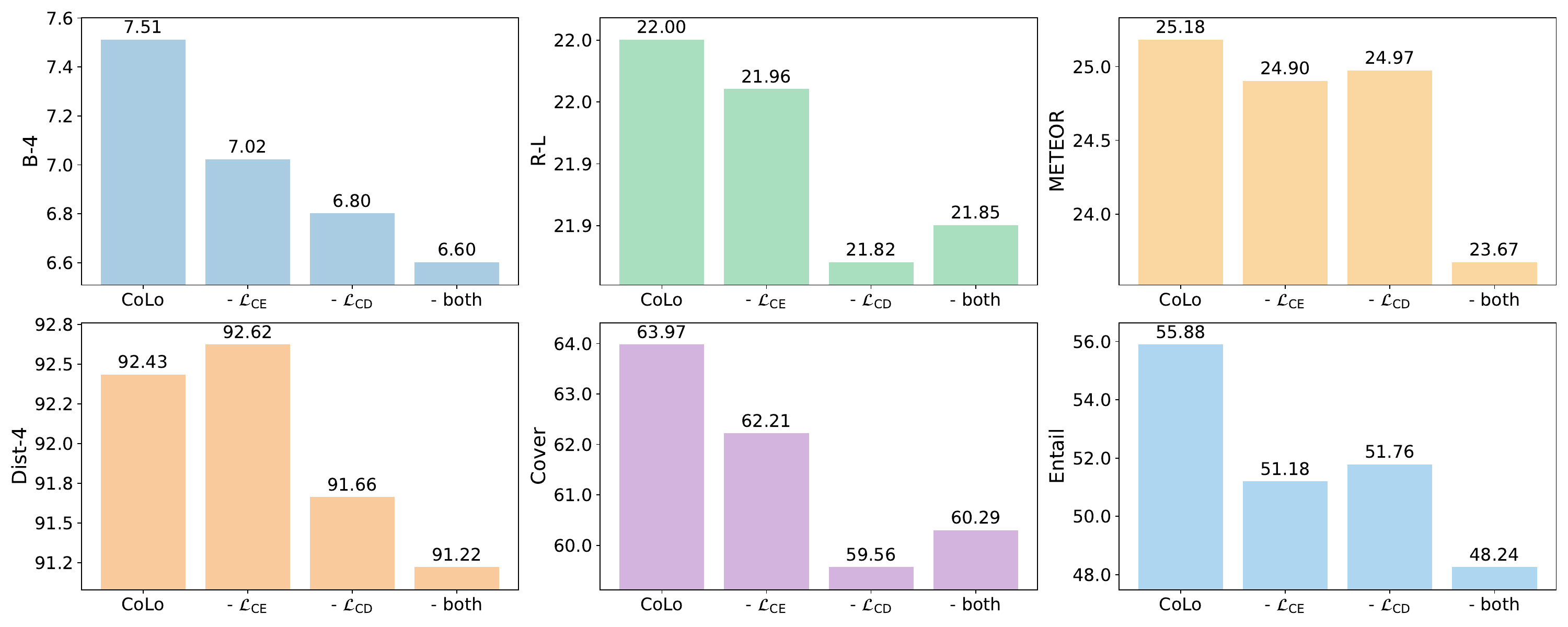}
\vspace{-6mm}
\caption{Results of ablation studies. - $\mathcal{L}_{\mathrm{CE}}$ excludes $\mathcal{L}_{\mathrm{CE}}$ from Colo. - $\mathcal{L}_{\mathrm{CD}}$ removes $\mathcal{L}_{\mathrm{CD}}$ from Colo. - both removes $\mathcal{L}_{\mathrm{CE}}$ and $\mathcal{L}_{\mathrm{CD}}$ from Colo.}
\label{figure:ablation_plot}
\end{center}
\vspace{-4mm}
\end{figure*}

\begin{table}[t!]
\caption{Effect of contrastive samples. The best results are bolded, and the runners-up are underlined.}
\label{tab:effect_of_contras}
\vspace{-2mm}
\centering
\footnotesize
\begin{tabular}{lccccccccc}
\toprule
\textbf{}     & $x_n^\mathrm{ES}$   & $x_n^\mathrm{AS}$    & $x_n^\mathrm{OS}$   & B-4 & R-L  & METEOR & Dist-4   & Cover   & Entail  \\ \midrule
CoLo  &\Checkmark&\Checkmark &\Checkmark& \textbf{7.51}  & \textbf{22.00}    & \textbf{25.18}  & \textbf{92.43}    & \textbf{63.97}   & \textbf{55.88}   \\
CoLo$_{e}$  &\Checkmark&           &          & \underline{7.20}  & 21.61    & \underline{24.62}  & \underline{91.49}    & \underline{61.32}   & \underline{51.76}   \\
CoLo$_{a}$  &          &\Checkmark &          & 6.54  & \underline{21.68}    & 23.91  & 90.99    & 58.68   & 48.24   \\
CoLo$_{o}$  &          &           &\Checkmark& 6.96  & 21.64    & 24.17  & 90.85    & 57.65   & 48.82  \\ \bottomrule
\end{tabular}
\vspace{-4mm}
\end{table}

To verify the effectiveness of our proposed Contrastive Encoding ($\mathrm{CE}$) and Contrastive Decoding ($\mathrm{CD}$), we conduct ablation studies by removing $\mathrm{CE}$, $\mathrm{CD}$ and both of them from our model (Figure \ref{figure:ablation_plot}). 
Our findings highlight the significant role played by both $\mathrm{CE}$ and $\mathrm{CD}$ in modeling \textit{comparative logical relations}. Removing either $\mathrm{CE}$ or $\mathrm{CD}$ results in a substantial decrease in performance. Furthermore, when both components are removed, the results decrease dramatically especially for the Cover and Entail metrics. All the findings validate the effectiveness of our method that learns to understand and generate text with \textit{comparative logical relations} in two stages (encoding and decoding). 

\begin{figure*}[t]
\vspace{-2mm}
\begin{center}
\includegraphics[width=1.0\textwidth]{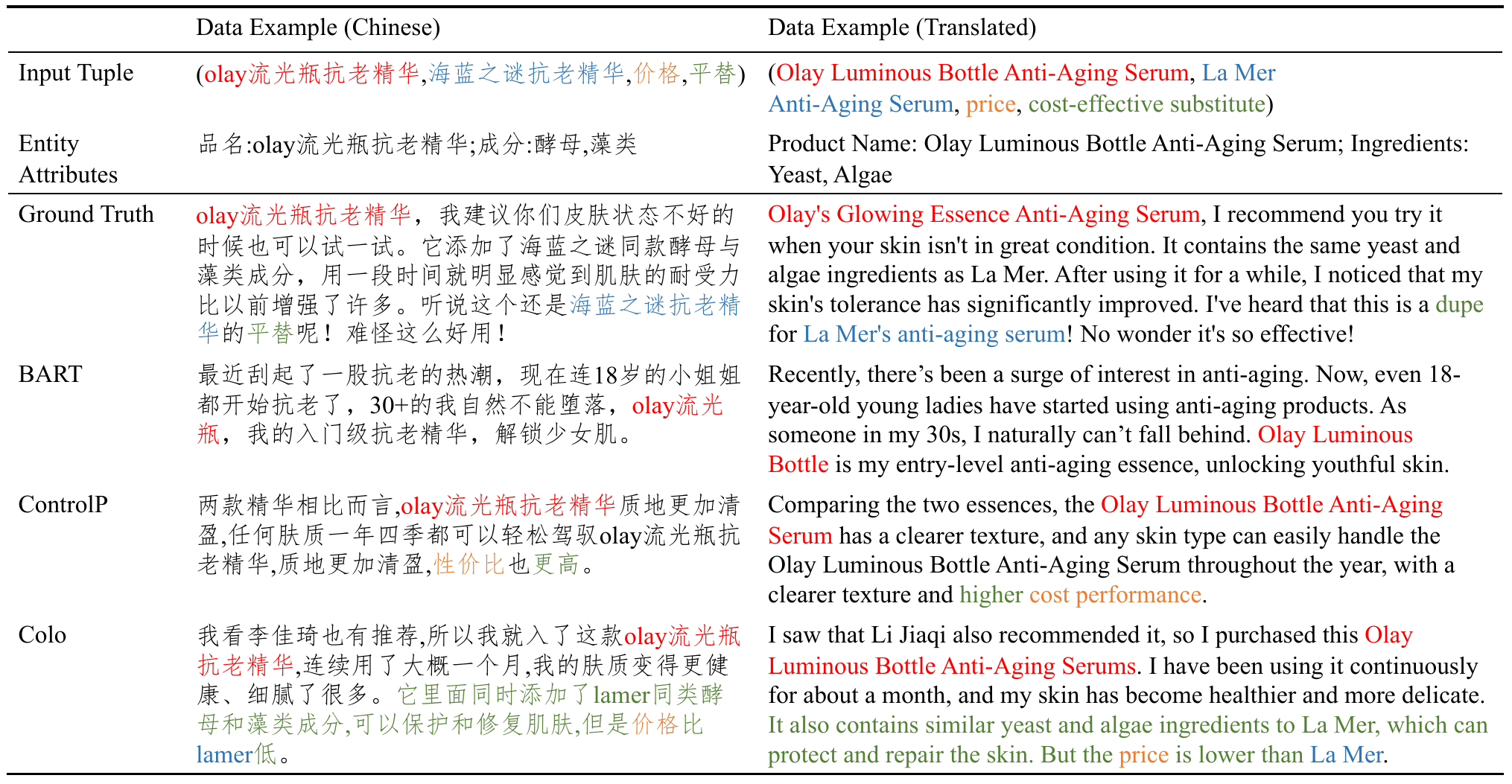}
\vspace{-6mm}
\captionof{table}{Descriptions generated by our model and baselines.}
\label{tab:case_study}
\end{center}
\vspace{-4mm}
\end{figure*}

\vspace{-2mm}
\subsection{Effect of Contrastive Samples}

To investigate the effectiveness of contrastive samples introduced in Section \ref{sec:cec}, we train CoLo with one category at a time (Table \ref{tab:effect_of_contras}). 
It is observed that when entities are swapped (CoLo$_{e}$), the model achieves the highest Entail compared to using other two types of contrastive samples. This finding supports our hypothesis stated in Section \ref{sec:cec} that pre-trained models are insensitive to the order of entities in the input sentence. Therefore, using the entity swapping based contrastive samples are more effective to boost the understanding and generation of text with CLRs. Furthermore, when all three types of contrastive samples are involved, the model achieves the best performance across all metrics, and the improvements are more significant regarding to the Cover and Entail metric. This outcome further validates the effectiveness of our strategy for constructing contrastive samples.

\vspace{-2mm}
\subsection{Case Studies}
\vspace{-1mm}
To have an intuitive understanding of the effectiveness of our method, we further analyze the text generated by the baselines and ours in Table \ref{tab:case_study}. To ensure a fair comparison, all models have approximately the same number of parameters.
We observe that the BART model merely focuses on describing the single entity as Olay Luminous Bottle, while completely neglecting the comparative relations and some entity attributes such as ingredients. Though the ControlP model generates some descriptions (``higher cost performance") about the relation, it is not complete for the absence of the compared entity. In addition, the attribute as the ingredient is also missing. In contrast, our CoLo model generates text with complete and accurate \textit{comparative logical relations}. Moreover, the descriptions cover all the attributes (green section in Table \ref{tab:case_study}), which provide well-founded and detailed explanations for the relation.

\vspace{-2mm}
\section{Related Work}
\vspace{-2mm}
\subsection{Data-to-Text Generation}
Data-to-Text Generation (D2T) is a Natural Language Generation (NLG) task that realizes the surface form of a generation from structured input data, such as spreadsheets \cite{parikh-etal-2020-totto} or keywords \cite{shao-etal-2019-long}.
Existing works mainly focus on how to generate high-quality descriptions with high fidelity, good coherence, and rich information for the entity itself. \cite{chan-etal-2019-stick,Chen2019TowardsKP,dan2024p,chan-etal-2020-selection}
Chan et al. \cite{chan-etal-2019-stick} proposed a Seq2Seq model with keyword memory considering both keywords and entity labels to ensure the high fidelity of generated descriptions. 
Chen et al. \cite{Chen2019TowardsKP} proposed a transformer-based model to generate personalized high-quality descriptions for a single product.
The most relevant work to ours is Chan et al. \cite{chan-etal-2020-selection}, which generated a description for a multi-product advertisement with a multi-agent framework.
They concentrated on selecting the most relevant entities to describe under a predefined topic with \textit{associative relations}.
Differently, we focus on modeling the \textit{comparative logical relations} between entities to generate high-quality descriptions. 

\subsection{Contrastive Learning}
Contrastive learning has been widely adopted in many natural language processing tasks. 
For instance, Das et al. \cite{das-etal-2022-container} utilized contrastive learning to optimize inter-token distribution distance for few-shot named entity recognition. 
Su et al. \cite{su-etal-2022-tacl} proposed a token-level contrastive loss to enhance the diversity of generated content.
These studies indicate that contrastive learning can improve the quality of generated text by enhancing the embeddings.
In this paper, we utilize contrastive learning to enhance the embeddings of the \textit{comparative logical relations} between two entities, which not only deepens the model's understanding of these relations, but also aids in generating more accurate descriptions.

\section{Conclusions}
In this paper, we propose an method to model the \textit{comparative logical relations} (CLRs) with two-staged contrastive learning for text generation, where the contrastive encoding facilitates the understanding of CLRs and the contrastive decoding forces to generate text with correct comparative logic. Extensive experiments have verified the effectiveness of our method by both automatic and human evaluations. Moreover, we provide a labeled Chinese Comparative Logical Relation Dataset (CLRD), which can help promote the research of text generation with multiple entities and fine-grained \textit{comparative logical relations}. In the future, we would like to investigate how to generate text following more complex logical relations. In addition, we will explore how to construct more effective contrastive samples to facilitate the understanding and generation of text. 

\section{Acknowledgement}
This research is funded by the National Science and Technology Major Project (No. 2021ZD0114002), the National Natural Science Foundation of China (No. 62307028), and the Science and Technology Commission of Shanghai Municipality Grant (No. 22511105901, No. 21511100402).

\bibliographystyle{splncs04}
\bibliography{mypaper}
%




\end{document}